\RequirePackage{fix-cm}
\documentclass{svjour3}                     
\smartqed  
\usepackage{graphicx}
\usepackage{latexsym}
\newcommand{\db}{D\!B}
\newcommand{\pdb}{p_{D\!B}}
\newcommand{\qdb}{q_{D\!B}}

\newcommand{\btheta}{\mbox{\boldmath $\theta$}}
\newcommand{\blambda}{\mbox{\boldmath $\lambda$}}

\newcommand{\bx}{\mbox{\boldmath $x$}}
\newcommand{\by}{\mbox{\boldmath $y$}}

\begin{document}

\title{A Logic-based  Approach to Generatively Defined Discriminative Modeling
}

\author{Taisuke Sato \and  Keiichi Kubota \and Yoshitaka Kameya}


\institute{Taisuke Sato \at
             Tokyo Institute of Technology, Japan \\
             Tel.: +81-3-5743-2186\\
              \email{sato@mi.cs.titech.ac.jp}           
      \and
           Keiichi Kubota \at
            Tokyo Institute of Technology, Japan \\
            Tel.: +81-3-5743-2186\\
           \email{kubota@mi.cs.titech.ac.jp}           
      \and
           Yoshitaka Kameya \at
             Meijo University, Japan \\
            Tel. : +81-52-838-2567\\
           \email{ykameya@meijo-u.ac.jp}           
}


\maketitle

\begin{abstract}
Conditional random fields (CRFs) are usually specified by graphical models but
in this paper we propose to  use probabilistic logic programs and specify them
generatively.  Our  intension is first to  provide a unified  approach to CRFs
for complex modeling through the use  of a Turing complete language and second
to  offer a  convenient way  of realizing  generative-discriminative  pairs in
machine learning  to compare generative  and discriminative models  and choose
the  best model.   We  implemented our  approach  as the  D-PRISM language  by
modifying PRISM, a logic-based  probabilistic modeling language for generative
modeling,  while exploiting  its dynamic  programming mechanism  for efficient
probability  computation.   We  tested  D-PRISM with  logistic  regression,  a
linear-chain  CRF and  a  CRF-CFG and  empirically  confirmed their  excellent
discriminative   performance  compared   to  their   generative  counterparts,
i.e.\ naive  Bayes, an  HMM and a  PCFG.  We  also introduced new  CRF models,
CRF-BNCs and CRF-LCGs.  They are  CRF versions of Bayesian network classifiers
and probabilistic  left-corner grammars respectively  and easily implementable
in  D-PRISM.  We  empirically  showed that  they  outperform their  generative
counterparts as expected.  
\end{abstract}

\section{Introduction}
Conditional  random fields (CRFs)  \cite{Lafferty01} are  probabilistic models
for discriminative modeling defining  a conditional distribution $p(y \mid x)$
over output $y$ given input $x$.  They are quite popular for labeling sequence
data  such as text  data and  biological sequences  \cite{Sutton12}.  Although
they  are  usually specified  by  graphical models,  we  here  propose to  use
probabilistic logic programs and  specify them generatively.  Our intension is
first to provide  a unified approach to CRFs for  complex modeling through the
use of  a Turing  complete language and  second to  offer a convenient  way of
realizing generative-discriminative  pairs \cite{Ng01} in  machine learning to
compare generative and discriminative models and choose the best model.

The use of logical  expressions to specify CRFs is not new  but they have been
used solely  as feature functions  \cite{Gutmann06,Richardson06}.  For example
in Markov logic networks  (MLNs)\cite{Richardson06}, weighted clauses are used
as  feature  functions  to  define  (conditional)  Markov  random  fields  and
probabilities are  obtained by Gibbs sampling.
In contrast,  our approach  is implemented by  a generative  modeling language
PRISM  \cite{Sato97,Sato01g}  where  clauses  have  no  weights;  they  simply
constitute a logic program $\db$  computing possible output $y$ from input $x$
by  proving  a  top-goal $G_{x,y}$  that  relates  $x$  to $y$.   In  addition
probabilities  are exactly  computed  by dynamic  programming.  $\db$  however
contains  special   atoms  of  the   form  ${\tt  msw}(i,v)$   having  weights
$\exp(\lambda_{i,v})$ where $i$ and $v$  are arbitrary terms.  They are called
${\tt  msw}$ atoms  here as  in PRISM.   We define  the weight  $q(x,y)$  of a
top-goal  $G_{x,y}$ as  a sum-product  of such  weights associated  with ${\tt
  msw}$ atoms  appearing in a proof  of $G_{x,y}$ and consider  $q(x,y)$ as an
unnormalized distribution.  By modifying  the dynamic programming mechanism of
PRISM slightly,  we can efficiently  compute, when possible and  feasible, the
unnormalized marginal  distribution $q(x)  = \sum_y q(x,y)$  and obtain  a CRF
$p(y \mid  x) = { q(x,y)  }/{ q(x) }$.   We implemented our idea  by modifying
PRISM  and  termed  the  resulting language  D-PRISM  (discriminative  PRISM).
D-PRISM is a  general programming language that generatively  defines CRFs and
provides built-in  predicates for parameter learning and  Viterbi inference of
CRFs.

Our approach  to CRFs  is general  in the sense  that, like  other statistical
relational learning  (SRL) languages for  CRFs \cite{Richardson06,McCallum09},
programs in D-PRISM have no restriction such as the exclusiveness condition in
PRISM \cite{Sato01g}  except for the use  of binary features and  we can write
any program, i.e.\  we can write arbitrary CRFs as long  as they are described
by D-PRISM. We point out that binary features are the most common features and
they can  encode basic  CRF models such  as logistic  regression, linear-chain
CRFs and CRF-CFGs  \cite{Johnson01,Finkel08,Sutton12}.  Furthermore by dynamic
programming, probabilistic  inference can be efficiently carried  out with the
same  time  complexity as  their  generative  counterparts  as exemplified  by
linear-chain CRFs and hidden Markov models (HMMs).

In machine learning it is  well-known that naive Bayes and logistic regression
form a  generative-discriminative pair \cite{Ng01}.  That  is, any conditional
distribution $p(y  \mid \bx)$ computed  from a joint distribution  $p(\bx,y) =
p(\bx \mid y)p(y)$  defined generatively by naive Bayes, where  $y$ is a class
and  $\bx$ is  a feature  vector,  can also  be defined  directly by  logistic
regression  and vice versa.   As is  empirically demonstrated  in \cite{Ng01},
classification accuracy  by discriminative models such  as logistic regression
is generally better  than their corresponding generative models  such as naive
Bayes  when  there is  enough  data but  generative  models  reach their  best
performance  more  quickly than  discriminative  ones  w.r.t.\  the amount  of
available data.  Also the theoretical analysis in \cite{Liang08} suggests that
when a model is wrong  in generative modeling, the deterioration of prediction
accuracy is  more severe than  in discriminative modeling. It  seems therefore
reasonable  to say  ``...For  any particular  data  set, it  is impossible  to
predict in advance whether a generative or a discriminative model will perform
better'' \cite{Sutton12}.   Hence what is  desirable is to provide  a modeling
environment in which the user can test both types of modeling smoothly without
pain and D-PRISM  provides such an environment that makes it  easy to test and
compare discriminative modeling and generative  modeling for the same class or
related family of probabilistic models.

In  what follows,  we review  CRFs in  Section~\ref{sec:crfs} and  also review
three basic models, i.e.\  logistic regression, linear-chain CRFs and CRF-CFGs
in    Section~\ref{sec:models}.      We    then    introduce     D-PRISM    in
Section~\ref{sec:d-prism}.   We empirically  verify the  effectiveness  of our
approach   in   Section~\ref{sec:experiments}    using   the   three   models.
Section~\ref{sec:new-models} introduces new CRF models, CRF-BNCs and CRF-LCGs,
both easily implementable in  D-PRISM.  In Section~\ref{sec:trans}, we discuss
program transformation  which derives a  program for incomplete data  from one
for  complete data.   Section~\ref{sec:discussion} contains  related  work and
discussion and Section~\ref{sec:conclusion} is the conclusion.

\section{Conditional random  fields}
\label{sec:crfs}

Conditional random  fields (CRFs) \cite{Lafferty01}  are popular probabilistic
models defining a  conditional distribution $p(\by \mid \bx)$  over the output
sequence  $\by$  given an  input  sequence  $\bx$  which takes  the  following
form\footnote{
Bold italic letters are (values of) random  vectors in this paper.
}:

\begin{eqnarray*}
p(\by \mid \bx) 
   & \equiv & \frac{1}{Z(\bx)}\exp\biggl\{ \sum_{k=1}^{K}\lambda_{k}f_k(\bx,\by) \biggr\}.
\end{eqnarray*}
Here $f_k(\bx,\by)$ and  $\lambda_{k}$ ($1 \leq k \leq  K$) are respectively a
real valued function ({\em feature  function}) and the associated weight ({\em
  parameter}) and $Z(\bx)$ a normalizing  constant.  As $Z(\bx)$ is the sum of
exponentially many  terms, the exact computation is  generally intractable and
takes $\mathcal{O}(M^{|{\boldmath y}|})$ time  where $M$ is the maximum number
of possible values for each component of $\by$ and hence approximation methods
have  been developed  \cite{Sutton12}.   However when  $p(\by  \mid \bx)$  has
recursive structure  of specific type  as a graphical model  like linear-chain
CRFs, $Z(\bx)$ is efficiently computable by dynamic programming.

Now let $D = \{ ({\bx}^{(1)},{\by}^{(1)}),\ldots,({\bx}^{(T)},{\by}^{(T)}) \}$
be  a training set.  The regularised  conditional log-likelihood  $l (\blambda
\mid D)$ of $D$ is given by

\begin{eqnarray*}
l (\blambda \mid D) 
& \equiv &
      \sum_{t=1}^{T}\log p(\by^{(t)} \mid \bx^{(t)})
         - \frac{\mu}{2}\sum_{k=1}^{K}\lambda_k^2 \nonumber\\
& = & \sum_{t=1}^{T}\biggl\{\sum_{k=1}^{K}\lambda_{k}f_k(\bx^{(t)},\by^{(t)})
         - \log Z(\bx^{(t)})\biggr\} - \frac{\mu}{2}\sum_{k=1}^{K}\lambda_k^2 
\end{eqnarray*}
where   $\blambda  =   \lambda_{1},\ldots,\lambda_{K}$   are  parameters   and
$\displaystyle{ \frac{\mu}{2}\sum_{k=1}^{K}\lambda_k^2 }\/$ is a penalty term.
Parameters are then estimated as the  ones that maximize $l (\blambda \mid D)$
by  Newton's  method  or  quasi-Newton  methods.  The  gradient  required  for
parameter learning is computed as

\begin{eqnarray*}
\frac{\partial l(\blambda \mid D)}{\partial\lambda_k} 
& = &
  \sum_{t=1}^{T}\biggl\{f_k(\bx^{(t)},\by^{(t)}) - E(f_k \mid \bx^{(t)})\biggr\}-\mu\lambda_k.
\end{eqnarray*}

The  problem  here  is  that  the expectation  $E(f_k  \mid  {\bx}^{(t)})$  is
difficult  to compute and  hence a  variety of  approximation methods  such as
stochastic gradient descent (SDG) \cite{Sutton12} have been proposed.  However
in this paper we focus on cases where exact computation by dynamic programming
is  possible  and  use   an  algorithm  that  generalizes  inside  probability
computation in probabilistic context free grammars (PCFGs) \cite{Manning99}.

After parameter learning, we apply our model to prediction tasks and infer the
most-likely output $\hat{\by}$  for an input sequence $\bx$  using
\begin{eqnarray*}
\hat{\by}
  & \equiv & \mbox{argmax}_{\boldmath y}p(\by \mid \bx)  \\
  & = &  \mbox{argmax}_{\boldmath y}\frac{1}{Z(\bx)}
          \exp\biggl\{\sum_{k=1}^{K}\lambda_{k}f_k(\bx,\by)\biggr\} 
 \; = \; \mbox{argmax}_{\boldmath y}\sum_{k=1}^{K}\lambda_k f_k(\bx,\by).
\end{eqnarray*}

As naively computing  $\hat{\by}$ is straightforward but too  costly, we again
consider only cases where dynamic  programming is feasible and apply a variant
of the Viterbi algorithm for HMMs.

\section{Basic models}
\label{sec:models}

\subsection{Logistic regression}
Logistic regression specifies a  conditional distribution $p(y \mid \bx)$ over
a  class variable  $y$ given  the input  $\bx =  x_1,\ldots,x_K$, a  vector of
attributes.  It assumes $\log p(y \mid \bx)$ is a linear function of $\bx$ and
given by

\begin{eqnarray*}
p(y \mid \bx)
  & \equiv & \frac{1}{Z(\bx)}\exp\biggl\{ \lambda_y+\sum_{j=1}^{K}\lambda_{y,j} x_j \biggr\}.
\end{eqnarray*}

We  here confirm  that logistic  regression is  a CRF.   Rewrite  $\lambda_y =
\sum_{y'}  \lambda_{y'}{\bf 1}_{\{y'=y\}}$  and $\lambda_{y,j}x_j  = \sum_{y'}
\lambda_{y',j}{\bf 1}_{\{y'=y\}}x_j$ and substitute  them for $\lambda_y $ and
$\lambda_{y,j}x_j$ in the above formula\footnote{
${\bf  1}_{\{y'=y\}}$  is a  binary  function of  $y$  taking  $1$ if  $y=y'$,
otherwise $0$.
}.  We obtain

\begin{eqnarray*}
p(y \mid \bx) 
  & =  &  \frac{1}{Z({\bx})}
            \exp\biggl\{ \sum_{y'} \lambda_{y'}{\bf  1}_{\{y'=y\}}
          + \sum_{y'}\sum_{j=1}^{K} \lambda_{y',j}{\bf 1}_{\{y'=y\}}x_j \biggr\}.
\end{eqnarray*}
By considering  ${\bf 1}_{\{y'=y\}}$ and  ${\bf 1}_{\{y'=y\}}x_j $  as feature
functions (of $y$ and $\bx$), we can see logistic regression is a CRF.

\subsection{Linear-chain CRFs}
CRFs   \cite{Lafferty01}  are   generally   intractable  and   a  variety   of
approximation  methods such  as sampling  and  loopy BP  have been  developed.
There  is  however  a tractable  subclass called  {\em  linear-chain  CRF\/}s.
They are of the following  the form:

\begin{eqnarray*}
p(\by \mid \bx)
  & \equiv & \frac{1}{Z(\bx)}
               \exp\biggl\{\sum_{k=1}^{K}\lambda_{k}\sum_{i=2}^{N} f_k(\bx,y_i,y_{i-1}) \biggr\}
\end{eqnarray*}
where $Z(\bx)$ is a normalizing constant.  They define, as CRFs, a conditional
distribution  $p(\by \mid  \bx)$ over  output sequences  $\by$ given  an input
sequence $\bx$  such that $|\bx| = |\by|  = N$ ($|\bx|$ denotes  the length of
vector   $\bx$)   but  feature   functions   are   restricted   to  the   form
$f(\bx,y_i,y_{i-1})$ ($\by  = y_1,\ldots,y_{N}, 2  \leq i \leq N$)  which only
refers to two consecutive components in $\by$.  Thanks to this local reference
restriction exact probability computation is possible for linear-chain CRFs in
time linear in  the input length $|\bx|$ by a  variant of the forward-backward
algorithm for  HMMs.  Linear-chain  CRFs are considered  as a  generalized and
undirected version of HMMs which enable us to use far richer feature functions
other than transition probabilities and emission probabilities used in HMMs.

\subsection{CRF-CFGs}
PCFGs \cite{Manning99}  are a basic class of  probabilistic grammars extending
CFGs by  assigning selection probabilities $\btheta$ to  production rules.  In
PCFGs, the  probability of  a sentence  is the sum  of probabilities  of parse
trees and  the probability  of a  parse tree is  the product  of probabilities
associated  with production  rules used  in  the tree.   PCFGs are  generative
models  and parameters are  usually learned  by maximum  likelihood estimation
(MLE).   So given  parse  trees $\tau_1,\ldots,\tau_T$  and the  corresponding
sentences   $s_1,\ldots,s_T$,  parameters  are   estimated  as   $\btheta^*  =
\mbox{argmax}_{\boldmath \theta} \prod_{t=1}^T p(\tau_t,s_t \mid \btheta)$.

Seeking  better  parsing accuracy,  Johnson  attempted  parameter learning  by
maximizing       conditional       likelihood:      $\btheta^{\dagger}       =
\mbox{argmax}_{\boldmath \theta} \prod_{t=1}^T p(\tau_t \mid s_t,\btheta)$ but
found  the  improvement  is  not statistically  significant  \cite{Johnson01}.
Later  Finkel et  al.\  generalized  PCFGs to  {\em  conditional random  field
  context  free grammar\/}s (CRF-CFGs)  \cite{Finkel08} where  the conditional
probability $p(\tau  \mid s)$ of a parse  tree $\tau$ given a  sentence $s$ is
defined by
\begin{eqnarray*}
p(\tau \mid s)
  & \equiv & 
    \frac{1}{Z(s)}\exp\biggl\{ \sum_{k=1}^{K}\lambda_k\sum_{r\in \tau}f_k(r,s) \biggr\}.
\end{eqnarray*}
Here  $Z(s)$  is  a  normalizing constant.   $\lambda_1,\ldots,\lambda_K$  are
parameters and  ${r \in  \tau}$ is  a CFG rule  (possibly enriched  with other
information) appearing  in the parse  tree $\tau$ of  $s$ and $f_k(r,s)$  is a
feature  function.   Finkel et  al.\  conducted  learning  experiments with  a
CRF-CFG using the Penn Treebank \cite{Marcus93}.  They learned parameters from
parse   trees   $\tau_1,\ldots,\tau_T$   and   the   corresponding   sentences
$s_1,\ldots,s_T$ in the corpus  by maximizing conditional likelihood just like
\cite{Johnson01} but  this time  they obtained a  significant gain  in parsing
accuracy \cite{Finkel08}.  Their experiments clearly demonstrate the advantage
of extensive use of features and discriminative parameter learning.

\section{D-PRISM}
\label{sec:d-prism}
Having seen basic models of CRFs, we next show how they are uniformly subsumed
by a  logic-based modeling language PRISM \cite{Sato97,Sato01g}  with a simple
modification of its probability  computation.  The modified language is termed
{\em D-PRISM\/} (discriminative PRISM).

\subsection{PRISM at a glance}
Before proceeding we quickly review PRISM\footnote{
{\tt http://sato-www.cs.titech.ac.jp/prism/}
}.   PRISM is  a  high-level  generative modeling  language  based on  Prolog,
extended  by a  rich array  of probabilistic  built-in predicates  for various
types  of probabilistic  inference  and parameter  learning.  Specifically  it
offers,  in  addition to  MLE  by the  EM  algorithm,  Viterbi training  (VT),
variational  Bayes  (VB),  variational   VT  (VB-VT)  and  MCMC  for  Bayesian
inference.    PRISM   has   been   applied   to   music   and   bioinformatics
\cite{Sneyers06,Biba11,Mork12}.

Syntactically a PRISM program $\db$ is  a Prolog program and runs like Prolog.
Fig.~\ref{prog:nb} is an  example of PRISM program for  naive Bayes.  $\db$ is
basically  a  set of  definite  clauses.   The  difference from  usual  Prolog
programs  is that  the  clause body  may  contain special  atoms, ${\tt  msw}$
atoms\footnote{
${\tt msw}$  stands for  ``multi-valued switch.''
}, of the form {\tt  msw($i$,$v$)} representing a probabilistic choice made by
(analogically speaking) rolling a die  $i$ and choosing the outcome $v$.  Here
$i$ is  a ground term  naming the ${\tt  msw}$ atom and  $v$ belongs to  a set
$V_i$  of  possible outcomes  declared  by  {\tt  values/2} declaration.   The
probability  of {\tt  msw($i$,$v$)} ($v  \in V_i$)  being true  is  denoted by
$\theta_{i,v}$  and  called  a   {\em  parameter\/}  for  {\tt  msw($i$,$v$)}.
Naturally  $\sum_{v  \in  V_i}  \theta_{i,v}  =  1$  holds.   Executing  ${\tt
  msw}(i,{\tt X})$  with a variable ${\tt X}$  returns a value $v  \in V_i$ in
${\tt  X}$  with  probability   $\theta_{i,v}$.   So  {\tt  msw(season,S)}  in
Fig.~\ref{prog:nb}  probabilistically  returns  one  of \{{\tt  spring},  {\tt
  summer}, {\tt fall}, {\tt winter}\} in {\tt S}.

\begin{figure}[th]
\rule{\textwidth}{0.25mm}\\ [-1em]
\begin{verbatim}
values(season,[spring,summer,fall,winter]).
values(attr(temp,_),[high,mild,low]).
values(attr(humidity,_),[high,low]).

nb([T,H],S):-                   % defines p(X,Y) where X = [T,H] and Y = S
    msw(season,S),              % choose S from {spring,summer,fall,winter}
    msw(attr(temp,S),T),        % choose T from {high,mild,low}
    msw(attr(humidity,S),H).    % choose H from {high,low}
nb([T,H]):- nb([T,H],_).        % defines p(X) where X = [T,H]
\end{verbatim}
\rule{\textwidth}{0.25mm}\\ 
\caption{ PRISM program $\db_0$ for naive Bayes }
\label{prog:nb}
\end{figure}

$\db$   defines   a    probability   measure   $\pdb(\cdot)$   over   Herbrand
interpretations (possible worlds)\cite{Sato01g}.  The probability $\pdb(G)$ of
a top-goal $G$  then is computed as a sum-product of  parameters in two steps.
First  $G$  is  reduced using  $\db$  by  SLD  search  to a  disjunction  $E_1
\vee\cdots\vee E_M$ such that each $E_j$  ($1 \leq j \leq M$) is a conjunction
of ${\tt msw}$ atoms representing  a sequence of probabilistic choices.  $E_j$
is called an  {\em explanation\/} for $G$ because it explains  why $G$ is true
or how  $G$ is proved as a  result of probabilistic choices  encoded by $E_j$.
Let $\phi(G) \equiv  \{ E_1,\ldots,E_M \}$ be the set  of all explanations for
$G$.  $\pdb(G)$  is computed as $\pdb(G)  = \sum_{E \in  \phi(G)} \pdb(E)$ and
$\pdb(E)   =   \prod_{k=1}^{N}   \theta_{k}$    for   $E   =   {\tt   msw}_{1}
\wedge\cdots\wedge{\tt msw}_{N}$, where $\theta_{k}$  is a parameter for ${\tt
  msw}_{k}$ ($1 \leq k \leq N$).

Let  $p(x,y)$ be  a joint  distribution over  input $x$  (or  observation) and
output $y$ (or hidden state) which we  wish to compute by a PRISM program.  We
write a program $\db$ that probabilistically proves $G_{x,y}$, a top-goal that
relates $x$  to $y$, using  ${\tt msw}$  atoms, in such  a way that  $p(x,y) =
\pdb(G_{x,y})$ holds.   Since $(x,y)$ forms complete data,  $G_{x,y}$ has only
one explanation $E_{x,y}$ for $G_{x,y}$\footnote{
This is an assumption but generally true with programs for complete data.
},  so  we  have  $p(x,y)   =  \pdb(G_{x,y})  =  \pdb(E_{x,y})  =  \prod_{i,v}
\theta_{i,v}^{\sigma_{i,v}(E_{x,y})}$  where  $\sigma_{i,v}(E_{x,y})$  is  the
count  of {\tt  msw($i$,$v$)}  in  $E_{x,y}$.  Introduce  $G_x  = \exists  y\,
G_{x,y}$.   Then the marginal  probability $p(x)$  is obtained  as $\pdb(G_x)$
because  $p(x) =  \sum_y p(x,y)  =  \sum_y \pdb(G_{x,y})  = \pdb(G_x)$  holds.
Hence the conditional distribution $p(y \mid x)$ is computed as
\begin{eqnarray}
p(y \mid x) &  = &
            {\displaystyle \frac{ \pdb(G_{x,y}) }{ \pdb(G_x) } }
    \;{=}\; {\displaystyle \frac{ \pdb(E_{x,y}) }{ \pdb(G_x) } }
    \;{=}\; {\displaystyle
       \frac{ \prod_{i,v} \theta_{i,v}^{\sigma_{i,v}(E_{x,y})} }
            { \sum_{E_{x,y} \in \phi(G_x)} \prod_{i,v} \theta_{i,v}^{\sigma_{i,v}(E_{x,y})} } }.
   \label{eq:crfprism}
\end{eqnarray}

We  next apply  (\ref{eq:crfprism})  to  the naive  Bayes  program $\db_0$  in
Fig.~\ref{prog:nb}.  $\db_0$  is intended  to  infer  a  season {\tt  S}  from
temperature  {\tt T} and  humidity {\tt  H} and  generatively defines  a joint
distribution    $p(\mbox{\tt    [T,H]},\mbox{\tt    S})    =    p_{\db_0}({\tt
  nb([T,H],S)})$.  To draw a sample from $p(\mbox{\tt [T,H]},\mbox{\tt S})$ or
equivalently  from  $p_{\db_0}(\mbox{\tt nb([T,H],S)})$,  it  first samples  a
season  {\tt S}  by executing  {\tt msw(season,S)},  then similarly  samples a
value {\tt T} of temperature and a value {\tt H} of humidity, each conditioned
on   {\tt    S},   by    executing   {\tt   msw(attr(temp,S),T)}    and   {\tt
  msw(attr(humidity,S),H)} in turn\footnote{
For  example {\tt msw(attr(temp,S),T)}  samples {\tt  T} from  the conditional
distribution $p({\tt T} \mid {\tt S})$.
}. Note that the program  also includes a clause {\tt nb([T,H]):-nb([T,H],\_)}
to compute  a marginal distribution  $p_{\db_0}( \mbox{\tt nb({[T,H]})}  )$ (=
$p(\mbox{\tt [T,H]})$).  The correspondence to (\ref{eq:crfprism}) is that $x$
= {\tt [T,H]}, $y$ = {\tt S}, $G_{x,y}$ = {\tt nb([T,H],S)} and $G_{x}$ = {\tt
  nb([T,H])}.   The conditional  distribution $p(\mbox{\tt  S}  \mid \mbox{\tt
  [T,H]})$  is  computed as  $p_{\db_0}(  \mbox{\tt nb([T,H],S)}  )/p_{\db_0}(
\mbox{\tt nb({[T,H]})} )$.
%
%

\subsection{From probability to weight}

The basic  idea of  our approach to  discriminative modeling is  to generalize
(\ref{eq:crfprism})   by  replacing   probability   $\theta_{i,v}$  for   {\tt
  msw($i$,$v$)}    with    arbitrary     {\em    weight\/}    $\eta_{i,v}    =
\exp(\lambda_{i,v})$.  In D-PRISM we further perform normalization to obtain a
CRF.  More precisely, we  first introduce an {\em unnormalized distribution\/}
$q(x,y) = \qdb(G_{x,y})$ defined by:
\begin{eqnarray*}
\qdb(G_{x,y})
  & \equiv &
    \exp\biggl( \sum_{i,v}\lambda_{i,v} \sigma_{i,v}(E_{\boldmath x,y}) \biggr) \\
  &    & \mbox{ where $E_{x,y}$ is a unique explanation for $G_{x,y}$}
\end{eqnarray*}
assuming that  for any  complete data $(x,y)$  and the  corresponding top-goal
$G_{x,y}$, our program, $\db$, always  has only one explanation $E_{x,y}$.  By
setting  $\lambda_{i,v} =  \ln  \theta_{i,v}$, $\qdb(G_{x,y})$  is reduced  to
$\pdb(G_{x,y})$ again.

Next  we rewrite  (\ref{eq:crfprism}) as  follows by  putting  $p(E_{x,y} \mid
G_{\boldmath  x})  \equiv  {\displaystyle  \frac{  \qdb(G_{x,y})  }{  \sum_{y}
    \qdb(G_{x,y}) }}$ and using $\eta_{i,v} = \exp(\lambda_{i,v})$.

\begin{eqnarray}
p(E_{x,y} \mid G_{\boldmath x})
  & = & \frac{1}{Z(G_{\boldmath x})}
            \exp\biggl( \sum_{i,v}\lambda_{i,v}\sigma_{i,v}(E_{\boldmath  x,y}) \biggr) 
  \; = \; \frac{1}{Z(G_{\boldmath x})}{ \prod_{i,v} \eta_{i,v}^{\sigma_{i,v}(E_{x,y})} }
                                      \label{eq:crf-def} \\
Z(G_{\boldmath x})
  & = & \sum_{E_{x,y} \in\phi(G_{\boldmath x})}
             \exp\biggl(\sum_{i,v}\lambda_{i,v}\sigma_{i,v}(E_{x,y})\biggr)   \nonumber \\
  & = & \sum_{E_{x,y} \in\phi(G_{\boldmath x})} \prod_{i,v} \eta_{i,v}^{\sigma_{i,v}(E_{x,y})}
                \label{eq:crf-def-Z}
\end{eqnarray}

(\ref{eq:crf-def})  and  (\ref{eq:crf-def-Z})  are fundamental  equations  for
D-PRISM describing how a CRF $p(y  \mid x) = p(E_{x,y} \mid G_{x})$ is defined
and  computed.   By comparing  (\ref{eq:crfprism})  to (\ref{eq:crf-def})  and
(\ref{eq:crf-def-Z}), we  notice that the most  computationally demanding task
in  D-PRISM, computing $Z(G_x)$  in (\ref{eq:crf-def-Z}),  can be  carried out
efficiently   by   dynamic   programming   just   by   replacing   probability
$\theta_{i,v}$ in PRISM  with weight $\eta_{i,v}$, resulting in  the same time
complexity for probability computation as in PRISM.

It is also  seen from (\ref{eq:crf-def}) and (\ref{eq:crf-def-Z})  that in our
formulation  of CRFs by  D-PRISM, $\sigma_{i,v}(E_{x,y})$,  the count  of {\tt
  msw($i$,$v$)} in  $E_{x,y}$, works as a (default) feature  function\footnote{
Since we assume that the top-goal $G_{x,y}$ has only one explanation $E_{x,y}$
for    a    complete     data    $(x,y)$,    $(x,y)$    uniquely    determines
$\sigma_{i,v}(E_{x,y})$.
} over the  input $x$ and  output $y$.  $\sigma_{i,v}(E_{x,y})$  becomes binary
when \mbox{\tt msw($i$,$v$)}  occurs at most once in  $E_{x,y}$.  For a binary
feature function  $f(x,y)$ in general,  let \mbox{\tt msw(f($x$,$y$),1)}  be a
dummy {\tt msw}  atom which is unique to $f(x,y)$ and  always true.  We assume
that corresponding to $f(x,y)$, there is a goal \mbox{\tt f($x$,$y$)} provable
in PRISM  if and only if $f(x,y)  = 1$.  Then it  is easy to see  that a PRISM
goal \mbox{\tt (f($x$,$y$) -> msw(f($x$,$y$),1) ; true)} realizes $f(x,y)$.

From  the viewpoint  of  modeling, we  emphasize  that for  the user,  D-PRISM
programs  are just  PRISM programs  that proves  two top-goals,  $G_{x,y}$ for
complete data $(x,y)$  and $G_{x}$ for incomplete data  $x$.  For example, the
PRISM program in Fig.~\ref{prog:nb} for  naive Bayes is also a D-PRISM program
defining logistic regression.\\

In  D-PRISM,  parameters  are  learned discriminatively  from  complete  data.
Consider the regularised (log) conditional likelihood $l (\blambda \mid D)$ of
a  set  of  observed data  $  D  =  \{  d_1,d_2,\ldots,d_T  \}$ where  $d_t  =
(G_{x^{(t)}},  E_{x^{(t)},y^{(t)}}) =  (G_t,E_t)$  ($1 \leq  t  \leq T$).   $l
(\blambda \mid D)$ is given by
\begin{eqnarray*}
l (\blambda \mid D)
  &\equiv &  \sum_{t=1}^{T}\log p(E_t \mid G_t) - \frac{\mu}{2}\sum_{i,v}\lambda_{i,v}^2 \\
  & =     &  \sum_{t=1}^{T}\biggl\{ \sum_{i,v}\lambda_{i,v}\sigma_{i,v}(E_t) - \log Z(G_t)
                          \biggr\} - \frac{\mu}{2}\sum_{i,v}\lambda_{i,v}^2
\end{eqnarray*}
and parameters  $\blambda =  \{ \lambda_{i,v} \}$  are determined as  the ones
that maximize $l (\blambda \mid  D)$.  Currently we use L-BFGS \cite{Liu89} to
maximize  $l (\blambda \mid  D)$.  The  gradient used  in the  maximization is
computed as
\begin{eqnarray}
 \frac{\partial l(\blambda|D)}{\partial\lambda_{i,v}}
  & = &
    \sum_{t=1}^{T}\biggl\{ \sigma_{i,v}(E_t)
              -\frac{\partial}{\partial\lambda_{i,v}}\log Z(G_t) \biggr\} - \mu\lambda_{i,v}
               \nonumber \\
   & =      &
     \sum_{t=1}^{T}\biggl\{ \sigma_{i,v}(E_t) - \sum_{E'\in\phi(G_t)}\sigma_{i,v}(E')p(E'\mid G_t)\biggr \}
        - \mu\lambda_{i,v}.               \label{eq:grad-d-prism}  \nonumber
\end{eqnarray}

Finally, Viterbi inference, computing the most likely output $y$ for the input
$x$, or the most likely  explanation $E_{x,y}^{\ast}$ for the top-goal $G_{x}$,
is  formulated as  (\ref{eq:crfviterbi}) in  D-PRISM and  computed  by dynamic
programming just like PRISM.
\begin{eqnarray}
E_{{\boldmath x},{\boldmath y}}^{\ast}
  & = & \mbox{argmax}_{ E_{x,y}\in\phi(G_{x}) }p(E_{x,y} \mid G_{x}) \nonumber \\
  & = & \mbox{argmax}_{ E_{x,y}\in\phi(G_{x}) }
              \frac{1}{Z(G_{x})}\exp\biggl( \sum_{i,v}\lambda_{i,v}\sigma_{i,v}(E_{x,y}) \biggr)  \nonumber \\
  & = & \mbox{argmax}_{ E_{x,y}\in\phi(G_{x}) }\sum_{i,v}\lambda_{i,v}\sigma_{i,v}(E_{x,y})
 \label{eq:crfviterbi}
\end{eqnarray}

\section{Experiments with three basic models}
\label{sec:experiments}

In this section, we conduct learning experiments with CRFs\footnote{
Experiments  in this paper  are done  on a  single machine  with Core  i7 Quad
2.67GHz$\times$2 CPU and 72GB RAM running OpenSUSE 11.2.
}.  CRFs are  encoded by D-PRISM programs while  their generative counterparts
are encoded  by PRISM programs.   We compare their accuracy  in discriminative
tasks.  We  consider three basic  models, logistic regression,  a linear-chain
CRF and a CRF-CFG, and learn their parameters by L-BFGS.

\subsection{Logistic regression with UCI datasets}

We select  four datasets with  no missing data  from the UCI  Machine Learning
Repository  \cite{Frank10} and  compare prediction  accuracy, one  by logistic
regression  written in  D-PRISM and  the  other by  a naive  Bayes model  (NB)
written in PRISM.

We use the  program in Fig.~\ref{prog:nb} with an  appropriate modification of
{\tt values/2} declarations.  The result by ten-fold cross-validation is shown
in   Table~\ref{tab:uci-nb}   with    standard   deviation   in   parentheses.
Table~\ref{tab:uci-nb-time} contains  learning time for each  dataset.  We can
see, except  for the zoo  dataset, logistic regression by  D-PRISM outperforms
naive Bayes by  PRISM at the cost of considerably  increased learning time for
larger datasets\footnote{
In this  paper, accuracies  in bold  letters indicate that  they are  the best
performance and the difference is  statistically significant by t-test at 0.05
significance level. Learning time is an average over five runs.
}.

\begin{table}[th]
\caption{\label{tab:uci-nb} Logistic-regression and naive Bayes : UCI datasets and accuracy}
\begin{center}
\begin{tabular}{|c||c|c|c||c|c|}
\hline
           \multicolumn{4}{|c||}{}   & D-PRISM              & PRISM \\ \hline\hline
Model    & \multicolumn{3}{|c||}{}   & logistic-regression  & naive Bayes   \\ \hline
Dataset  & Size  & \#Class & \#Attr. &    \multicolumn{2}{|c|}{}            \\ \hline
zoo      & 101   & 7       & 16      & 96.00\%(5.16)        & 97.0\%(6.74) \\ \hline
car      & 1728  & 4       & 6       & {\bf 93.28}\%(2.02)  & 86.11\%(1.47) \\ \hline
kr-vs-kp & 3196  & 2       & 36      & {\bf 93.58}\%(4.40)  & 87.92\%(1.69) \\ \hline
nursery  & 12960 & 5       & 8       & {\bf 92.54}\%(0.60)  & 90.27\%(0.97) \\ \hline
\end{tabular}
\end{center}
\mbox{}\\  
\caption{\label{tab:uci-nb-time} Logistic-regression and naive Bayes : Learning time (sec) }
\begin{center}
\begin{tabular}{|c||c|c|c|}
\hline
         &  D-PRISM              & PRISM       \\ \hline\hline
Model    &  logistic-regression  & naive Bayes \\ \hline
Method   &  L-BFGS               & counting    \\ \hline
zoo      &  0.09(0.00)           & 0.04(0.00)  \\ \hline
car      &  0.04(0.00)           & 0.04(0.00)  \\ \hline
kr-vs-kp &  145.35(0.41)         & 0.30(0.00)  \\ \hline
nursery  &  321.65(1.23)         & 0.38(0.00)  \\ \hline
\end{tabular}
\end{center}
\end{table}

\subsection{Linear-chain CRF with the Penn Treebank} 

We here  compare a linear-chain  CRF encoded by  a D-PRISM program and  an HMM
encoded  by  a PRISM  program  using sequence  data  extracted  from the  Penn
Treebank \cite{Marcus93}.  What  we actually do is to write  an HMM program in
PRISM for complete  data and another program for  incomplete data and consider
their union as a D-PRISM program defining a linear-chain CRF, similarly to the
case of naive Bayes and logistic regression.  For simplicity we employ default
features, i.e.\ the count of various {\tt msw} atoms in an explanation.

\begin{figure}[b]
\rule{\textwidth}{0.25mm}\\ [-1em]
\begin{verbatim}
% HMM specification              % HMM specification for the Penn tree bank 
%      for a sample HMM          %      
values(init,[s0,s1]).            % values(init,[NNP,VBZ,s_dot,t_s_paren_l,...])
values(tr(_),[s0,s1]).           % values(tr(_),[NNP,VBZ,s_dot,t_s_paren_l,...]
values(out(_),[a,b]).            % values(out(_),[mss_dot,haag,plays,elianti,...]

hmm0([X0|Xs],[Y0|Ys]):- msw(init,Y0),msw(out(Y0),X0),hmm1(Y0,Xs,Ys).
hmm1(_,[],[]).
hmm1(Y0,[X|Xs],[Y|Ys]):- msw(tr(Y0),Y),msw(out(Y),X),hmm1(Y,Xs,Ys).

hmm0([X|Xs]):- msw(init,Y0),msw(out(Y0),X),hmm1(Y0,Xs).
hmm1(_,[]).
hmm1(Y0,[X|Xs]):- msw(tr(Y0),Y),msw(out(Y),X),hmm1(Y,Xs).
\end{verbatim}
\rule{\textwidth}{0.25mm}\\ 
\caption{Linear-chain CRF program}
\label{prog:crf-hmm}
\end{figure}

Fig.~\ref{prog:crf-hmm} is a sample D-PRISM  program for a CRF with two states
\{{\tt s0},  {\tt s1}\} and  two emission symbols  \{{\tt a}, {\tt  b}\}. {\tt
  hmm0/2}   describes  complete   data   and  corresponds   to  $G_{x,y}$   in
(\ref{eq:crfprism})  whereas   {\tt  hmm0/1}   is  for  incomplete   data  and
corresponds to $G_{x}$ in (\ref{eq:crfprism})\footnote{
Using  ``{\tt hmm0([X0|Xs]):-  hmm0([X0|Xs],\_)}'' to  define {\tt  hmm0/1} is
possible and theoretically correct but would kill the effect of tabling.  This
problem is discussed in Section~\ref{sec:trans}.
}. As a  CRF program, ground {\tt msw} atoms such  as {\tt msw(init,s0)}, {\tt
  msw(tr(s0),s1)} and  {\tt msw(out(s0,a))} represent  binary feature functions
over sequences  of state  transitions and emitted  symbols.  For  example {\tt
  msw(tr(s0),s1)} returns 1 (true) if the state transition sequence contains a
transition from {\tt s0} to {\tt s1} else 0 (false).

We conduct  a comparison of prediction  accuracy by a linear-chain  CRF and an
HMM  using the  D-PRISM program  in Fig.~\ref{prog:crf-hmm}.   The task  is to
predict the  POS (Part Of Speech)  tag sequence (hidden state  sequence) for a
given sentence (emitted  symbol sequence).  As learning data,  we use two sets
of pairs  of sentence and  POS tag sequence  extracted from the  Penn Treebank
\cite{Marcus93}: section-02  in the WSJ (Wall Street  Journal articles) corpus
referred  to  here  as WSJ02-ALL  and  its  subset  referred to  as  WSJ02-15,
consisting of data  of length less-than or equal to  15.  Their statistics are
shown in Table~\ref{tab:penn}.

Table~\ref{tab:crf-hmm-penn} contains  prediction accuracy (\%)  by eight-fold
cross-validation  and  learning  time  taken for  WSJ02-ALL.   Parameters  are
learned by  L-BFGS for D-PRISM and  by counting for PRISM.   The table clearly
demonstrates again  that we can  achieve better prediction performance  at the
cost of increased  learning time; D-PRISM gains 5.94\%  increase in prediction
accuracy for  the WSJ02-15 dataset but  learning time by L-BFGS  in D-PRISM is
about 60 times longer than that by counting in PRISM.

\begin{table}[h]
\caption{\label{tab:penn} Penn Treebank data}
\begin{center}
\begin{tabular}{|c||c|c|c|c|}
\hline
Dataset    & Size & Ave-len  & \#Tags & \#Words  \\ \hline\hline
WSJ02-15   & 1087 &   9.69   & 40     & 3341     \\ \hline
WSJ02-ALL  & 2419 &  19.28   & 45     & 8476     \\ \hline
\end{tabular}
\end{center}
\mbox{}\\
\caption{\label{tab:crf-hmm-penn} Linear-chain CRF and HMM : Labeling accuracy and learning time}
\begin{center}
\begin{tabular}{|cc||c|c|c|c|c|c|}
\hline
          &                                 & D-PRISM             & PRISM          \\ \hline\hline
\multicolumn{2}{|c||}{Model}                & linear-chain CRF    & HMM            \\ \hline
\multicolumn{2}{|c||}{Method}               & L-BFGS              & counting       \\ \hline
Accuracy & (WSJ02-15)                       & {\bf 83.17}\%(1.23) & 77.23\%(1.38)  \\ \cline{2-4}
         & (WSJ02-AL)                       & {\bf 90.60}\%(0.32) & 87.27\%(0.29)  \\ \hline
\multicolumn{2}{|c||}{Learning time (sec)}  &                     &  \\
\multicolumn{2}{|c||}{(WSJ02-15)}           & 499.34(1.06)        & 8.04 (0.00)    \\  \hline
\end{tabular}
\end{center}
\end{table}


%



\subsection{CRF-CFG with the ATR tree corpus}
\label{subsec:CRFCFG}

We here  deal with  probabilistic grammars which  graphical models  are unable
even to represent. We compare the parsing accuracy of a CRF-CFG described by a
D-PRISM program and that of a PCFG described by a PRISM program. We do not use
features other than the  count of a rule in the parsing  tree.  To save space,
we omit programs though they are (almost) identical.

As  a   dataset,  we  use  the   ATR  tree  corpus  and   its  associated  CFG
\cite{Uratani94}.   Their statistics  are shown  in Table~\ref{tab:atrcorpus}.
After parameter learning by regularised conditional likelihood for the CRF-CFG
and by the usual likelihood for the PCFG, we compare their parsing accuracy by
ten-fold  cross-validation.  The  task  is to  predict  a parse  tree given  a
sentence  and  the  predicted  tree  is considered  correct  when  it  exactly
coincides with the one for the  sentence in the ATR tree corpus (exact match).
As a reference, we also measure  parsing accuracy by PCFG whose parameters are
learned from incomplete data, i.e. sentences by the EM algorithm in PRISM.

\begin{table}[th]
\caption{ \label{tab:atrcorpus} ATR corpus }
\begin{center}
 \begin{tabular}{|c|c|c|c|c|}
 \hline
  Size &  Ave-len & \#Rules & \#Nonterminals & \#Terminals \\ \hline
 10995 &  9.97    & 861     &   168          & 446   \\  \hline
 \end{tabular}
\end{center}
\mbox{}\\ 
\caption{ \label{tab:atrcrf} CRF-CFG and PCFG : Parsing accuracy and learning time }
\begin{center}
 \begin{tabular}{|c||c|c|c|}
 \hline
                    &  D-PRISM             &  \multicolumn{2}{c|}{PRISM}    \\  \hline\hline
Model               &  CRF-CFG             &  \multicolumn{2}{c|}{PCFG}     \\  \hline
Method              &  L-BFGS              & counting      & EM             \\  \hline
Accuracy            & {\bf 82.74}\%(1.62)  & 79.06\%(1.25) & 70.02\%(0.87)   \\ \hline  
Learning time (sec)  & 205.51 (0.79)        & 2.30 (0.26)   & 65.72 (1.46)    \\ \hline  
\end{tabular}
\end{center}
\end{table}

Table~\ref{tab:atrcrf}  tells us  that when  a  tree corpus  is available,  as
reported in  \cite{Finkel08}, shifting from PCFG (PRISM)  to CRF-CFG (D-PRISM)
yields much better prediction performance (and shifting cost is almost zero if
we use D-PRISM) but at the same time this shifting incurs almost two orders of
magnitude longer learning time.

\section{ Exploring new models }
\label{sec:new-models}
In  this section,  we demonstrate  how the  power of  D-PRISM is  exploited to
explore new probabilistic models.  We propose two new models.  One is CRF-BNCs
which  are a  CRF  version of  Bayesian  networks classifiers.   The other  is
CRF-LCGs which  are a CRF  version of probabilistic left-corner  grammars that
generatively formalize probabilistic  left-corner parsing.  We first introduce
CRF-BNCs.

\subsection{ CRF-BNCs }
Bayesian   network   classifiers   (BNCs)  \cite{Friedman97,Cheng99}   are   a
generalization of naive Bayes  classifiers. They use general Bayesian networks
(BNs) as  a classifier  and allow dependencies  among attributes  unlike naive
Bayes classifiers.  Although BNCs outperform NBs classifiers in accuracy, they
are  still  generative.   We  here  introduce  a CRF  version  of  BNCs,  {\em
  conditional  random  field BNC\/}s  (CRF-BNCs),  and  empirically show  that
CRF-BNCs can outperform  BNCs.  Due to space limitations,  we explain CRF-BNCs
by an example.

\begin{figure}[h]
\centerline{\includegraphics[scale=0.45]{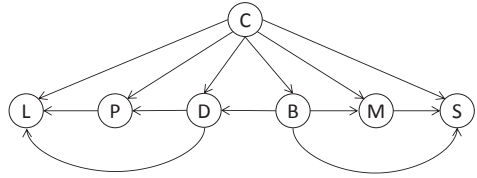}}
\caption{ Bayesian network for the car dataset }
\label{graph:bn-car}
\end{figure}

CRF-BNCs   are  obtained,  roughly   speaking,  by   generalizing  conditional
probability  tables in Bayesian  networks to  potential functions  followed by
normalization w.r.t.\ the class variable\footnote{
Since   CRF-BNCs  preserve   the   graph  structure   of  Bayesian   networks,
probabilistic inference  by belief propagation can be  efficiently carried for
both of them with the same time complexity.
}.   Fig.~\ref{graph:bn-car} is  an example  of Bayesian  network for  the car
dataset in the UCI Machine  Learning Repository \cite{Frank10}. It has a class
variable {\tt C} and six attribute  variables, {\tt B}, {\tt M}, {\tt D}, {\tt
  P}, {\tt  L} and {\tt  S}.  We assume  they have dependencies  designated in
Fig.~\ref{graph:bn-car}.

\begin{figure}[b]
\rule{\textwidth}{0.25mm}\\ [-1em]
\begin{verbatim}
values(class,[unacc,acc,good,vgood]).
values(attr(buying,_),[vhigh,high,med,low]).
...
values(attr(safety,_),[low,med,high]).

bn(Attrs):- bn(Attrs,_).           % defines q(x) where x = Attrs
bn(Attrs,C):-                      % defines q(x,y) where x = Attrs, y = C
   Attrs = [B,M,D,P,L,S],
   msw(class,C), msw(attr(buying,[C]),B), msw(attr(maint,[B,C]),M),
   msw(attr(doors,[B,C]),D), msw(attr(persons,[D,C]),P),
   msw(attr(lug_boot,[D,P,C]),L), msw(attr(safety,[B,M,C]),S).
\end{verbatim}
\rule{\textwidth}{0.25mm}\\ 
\caption{CRF-BN program $\db_1$ for the car dataset}
\label{prog:bn-car}
\end{figure}

Implementing  a CRF-BNC for  Fig.~\ref{graph:bn-car} is  easy in  D-PRISM.  We
have  only to write  a usual  generative Bayesian  network program  $\db_1$ in
PRISM shown  in Fig.~\ref{prog:bn-car}  and run it  as a D-PRISM  program.  In
Fig.~\ref{prog:bn-car}, the  first clause about {\tt bn}  predicate defines an
unnormalized probability $q_{\db_1}(\mbox{\tt  bn(Attrs)})$ and the second one
defines $q_{\db_1}(\mbox{\tt bn(Attrs,C)})$.   So the conditional distribution
$p({\tt   C}  \mid   {\tt   Attrs})$  is   computed  as   $q_{\db_1}(\mbox{\tt
  bn(Attrs,C)})/q_{\db_1}(\mbox{\tt bn(Attrs)})$
%

We conduct a learning experiment similarly to Section~\ref{sec:experiments} to
compare the CRF-BNC in Fig.~\ref{prog:bn-car}  and its original BNC and obtain
Table~\ref{tab:crf-uci}   for  accuracy   by  ten-fold   cross-validation  and
Table~\ref{tab:crf-uci-time} for learning time of each dataset\footnote{
Due to space  limitations Bayesian networks for {\tt  zoo}, {\tt kr-vs-kp} and
{\tt nursery} are omitted.%

}.  Our  experiment, though  small, strongly suggests  that when  datasets are
large enough,  CRF-BNCs can  outperform BNCs by  a considerable margin  at the
cost of long learning time.

\begin{table}[h]
\caption{\label{tab:crf-uci} CRF-BNC and BNC : UCI datasets and accuracy}
\begin{center}
\begin{tabular}{|c||c|c|c||c|c|}
\hline
            \multicolumn{4}{|c||}{}  & D-PRISM              & PRISM \\ \hline\hline
Model    &  \multicolumn{3}{|c||}{}  & CRF-BNC              & BNC   \\ \hline
Dataset  & Size  & \#Class & \#Attr. &                      &               \\ \hline
zoo      & 101   & 7       & 16      & 98.0 \%(4.21)        & 98.09\%(4.03) \\ \hline
car      & 1728  & 4       & 6       & {\bf 99.82}\%(0.54)  & 91.55\%(1.92) \\ \hline
kr-vs-kp & 3196  & 2       & 36      & {\bf 97.87}\%(0.85)  & 88.76\%(1.31) \\ \hline
nursery  & 12960 & 5       & 8       & {\bf 96.57}\%(0.43)  & 92.46\%(0.59) \\ \hline
\end{tabular}
\end{center}
\mbox{}\\
\caption{\label{tab:crf-uci-time} CRF-BNC and BNC : Learning time (sec) }
\begin{center}
\begin{tabular}{|c||c|c|c|}
\hline
         &  D-PRISM        & PRISM      \\ \hline\hline
Model    &  CRF-BNC        & BNC        \\ \hline
Method   &  L-BFGS         & counting   \\ \hline
zoo      &    0.12(0.00)   & 0.05(0.00) \\ \hline
car      &    0.92(0.00)   & 0.08(0.00) \\ \hline
kr-vs-kp &   53.58(4.15)   & 0.34(0.00) \\ \hline
nursery  &  106.65(6.71)   & 0.42(0.00) \\ \hline
\end{tabular}
\end{center}
\end{table}

\subsection{ CRF-LCGs }
A  second new  model  class is  {\em  CRF-left-corner grammar\/}s  (CRF-LCGs).
CRF-LCGs are a  CRF version of probabilistic left-corner  grammars (PLCGs) and
considered dual  to CRF-CFGs \cite{Finkel08} in  the sense that  the former is
based     on     bottom-up      parsing,     i.e.\     left-corner     parsing
\cite{Manning97,Uytsel01}   whereas   the   latter   is  based   on   top-down
parsing. Although left-corner parsing  is more context-dependent than top-down
parsing and accordingly CRF-LCGs are  expected to perform better than CRF-CFGs
in parsing, no proposal of CRF-LCGs has been made yet to our knowledge.

\begin{figure}[b]
\centerline{\includegraphics[scale=0.4]{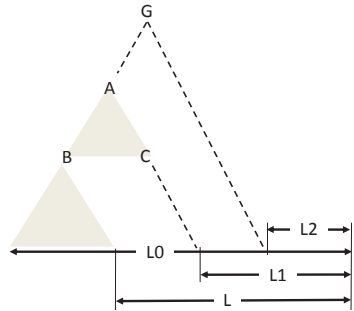}}
\caption{ Partial parse trees constructed in left-corner parsing }
\label{graph:plc-tree}
\end{figure}

\begin{figure}[hbt]
\rule{1.0\hsize}{0.8pt}
{\footnotesize
\begin{verbatim}
plcg(L0):-                                      plcg(L0,T):-
   start_symbol(C),                               start_symbol(C),
   g_call([C],L0,[]).                             g_call_t([C],L0,[],[T]).

g_call([],L,L).                                g_call_t([],L,L,[]).
g_call([G|R],[Wd|L],L2):-                      g_call_t([G|R],[Wd|L],L2,T):-
   ( terminal(G) ->                               ( terminal(G) ->
         G = Wd, L1 = L                               G = Wd, L1 = L, T = [Wd|TR]
   ; msw(first(G),Wd),                            ; msw(first(G),Wd), T = [TG|TR],
        lc_call(G,Wd,L,L1) ),                         lc_call_t(G,Wd,L,L1,Wd,TG) ),
   g_call(R,L1,L2).                            g_call_t(R,L1,L2,TR).

lc_call(G,B,L,L2):-                            ... 
   msw(lc(G,B),rule(A,[B|RHS2])),
   g_call(RHS2,L,L1),
   ( G == A -> attach_or_project(A,Op),
       ( Op == attach, L2=L1
       ; Op == project, lc_call(G,A,L1,L2) )
   ; lc_call(G,A,L1,L2) ).
attach_or_project(A,Op):-
   ( reachable(A,A) -> msw(attach(A),Op) ; Op = attach ).
\end{verbatim}
}
\rule{1.0\hsize}{0.8pt}
\caption{ PLCG parsers for sentences (left) and for sentence-tree pairs (right) }
\label{prog:plcg}
\end{figure}

Recall that left-corner parsing  is procedurally defined through three parsing
operations,  i.e.\ shift,  attach  and  project.  However  it  can be  defined
logically by a  pure logic program that describes  various relationships among
partial parse trees spanning substrings of the input sentence.  Let {\tt N} be
a nonterminal  and call a partial parse  tree with root {\tt  N} {\tt N}-tree.
Fig.~\ref{graph:plc-tree} is  a snapshot of left-corner parsing  when the {\tt
  B}-tree is projected by a CFG rule {\tt A -> B C} to complete a {\tt G}-tree
where   {\tt   G}   and   {\tt    A}   are   in   the   left-corner   relation
\cite{Manning97,Manning99}.

By translating the  relationships that hold among various  partial parse trees
in  Fig.~\ref{graph:plc-tree} into  a  logic program,  we  obtain a  bottom-up
parser for  probabilistic left-corner grammars  as illustrated on the  left in
Fig.~\ref{prog:plcg}. There, for  example, {\tt lc\_call(G,B,L,L2)} holds true
for the parsing  configuration described in Fig.~\ref{graph:plc-tree} (details
omitted)\cite{Manning97}.  Similarly  we write a parsing program  in PRISM for
complete data (sentence {\tt L0} and its  tree {\tt T}) placed on the right in
Fig.~\ref{prog:plcg},  which is almost  isomorphic to  the left  program.  The
left and right  PRISM programs combined together constitute  a D-PRISM program
for CRF-LCGs ({\tt values} declarations that specify CFG rules are not shown).

Following the  case of CRF-CFG  and PCFG in  Section~\ref{sec:experiments}, we
measure the parsing accuracy by  ten-fold cross-validation of CRF-LCG and PLCG
for  the ATR  corpus  and the  associated  CFG using  the  D-PRISM program  in
Fig.~\ref{prog:plcg}.   The   result  is  shown   in  Table~\ref{tab:crf-lcg}.
CRF-LCG  achieves the  highest parsing  accuracy  compared to  PLCG, PCFG  and
CRF-CFG but again at the cost of long learning time.

\begin{table}[h]
\caption{ \label{tab:crf-lcg} CRF-LCG and PLCG :  Parsing accuracy and learning time}
\begin{center}
 \begin{tabular}{|c||c|c|c|}
 \hline
                    &  D-PRISM             &  \multicolumn{2}{c|}{PRISM}     \\ \hline\hline
Model               &  CRF-LCG             &  \multicolumn{2}{c|}{PLCG}      \\ \hline
Method              &  L-BFGS              & counting      & EM              \\ \hline
Accuracy            & {\bf 87.26}\% (0.99) & 82.70\%(1.97) & 72.45\%(1.37)   \\ \hline    
Learning time (sec)  & 290.89 (1.86)        & 9.41 (0.15)   & 102.24 (8.61)   \\ \hline    
\end{tabular}
\end{center}
\end{table}

\section{ Program transformation for incomplete data }
\label{sec:trans}

As explained in Section~\ref{sec:d-prism}, in D-PRISM the user needs to define
two top-goals,  $G_{x,y}$ for complete  data $(x,y)$ and $G_x$  for incomplete
data  $x$,  each  defining  unnormalized distributions  $\qdb(G_{x,y})  $  and
$\qdb(G_{x})$ respectively.  Then $p(y \mid x)$ is computed as ${\displaystyle
  \frac{  \qdb(G_{x,y})  }{  \qdb(G_{x})  }}$.  However  since  $G_{x,y}$  and
$G_{x}$ are logically connected  as $G_{x} \Leftrightarrow \exists y G_{x,y}$,
it is theoretically enough and  correct to add a clause ``{\tt g(X):-g(X,Y)}''
to the program for $G_{x,y}$ to obtain a program for $\qdb(G_{x})$\footnote{
Here  it is  assumed that  ``{\tt g(X,Y)}''  is a  top-goal for  $G_{x,y}$ and
``{\tt g(X)}'' for $G_{x}$ respectively.
}.  This is what we did for the naive Bayes program in Fig.~\ref{prog:nb}.

Unfortunately this  simple approach does not  work in general.   The reason is
that the  search for  all explanations for  ${\tt g(X)}$ causes  an exhaustive
search for  proofs of  ${\tt g(X,Y)}$  for all possible  values of  ${\tt Y}$.
Consequently when a subgoal occurring in the search process that carries ${\tt
  Y}$ is proved with some value ${\tt  Y} = a$ and tabled, i.e.\ stored in the
memory for reuse, it has little chance of being reused later because ${\tt Y}$
in the subgoal mostly takes different values from $a$.  As a result the effect
of tabling  is almost  nullified, causing an  exponential search time  for all
explanations for ${\tt g(X)}$.

To avoid this  negative effect of the redundant argument,  ${\tt Y}$, we often
have  to write  a specialized  program for  ${\tt g(X)}$,  independently  of a
program for ${\tt g(X,Y)}$,  that does not refer to ${\tt Y}$;  in the case of
linear-chain CRF  program in  Fig.~\ref{prog:crf-hmm}, we wrote  two programs,
one for  ${\tt hmm0(X,Y)}$ (complete data)  and the other  for ${\tt hmm0(X)}$
(incomplete data).  The latter is a usual HMM program and efficient tabling is
possible that  guarantees linear time  search for all  explanations.  However,
writing  a specialized  program for  ${\tt g(X)}$  invites another  problem of
program correctness.  When we  independently write two programs, ${\db}_1$ for
${\tt g(X,Y)}$ and ${\db}_2$ for ${\tt g(X)}$, they do not necessarily satisfy
${\qdb}_1(\exists  {\tt  X} \mbox{\tt  g(X,Y)})  = {\qdb}_2(\mbox{\tt  g(X)})$
which is  required for  sound computation of  the conditional  distribution of
$p(y \mid  x)$.  It is  therefore hoped to  find a way of  obtaining ${\db}_2$
satisfying this property.

\begin{figure}[b]
\rule{\textwidth}{0.25mm}\\ [-1em]
\begin{verbatim}
 (1) hmm0([X0|Xs],[Y0|Ys]):- msw(init,Y0),msw(out(Y0),X0),hmm1(Y0,Xs,Ys).
 (2) hmm1(_,[],[]).
 (3) hmm1(Y0,[X|Xs],[Y|Ys]):- msw(tr(Y0),Y),msw(out(Y),X),hmm1(Y,Xs,Ys).
 (4) hmm0(X):- hmm0(X,Y).
 (5) hmm1(Y0,Xs):- hmm1(Y0,Xs,Ys).

 (6) hmm0([X0|Xs]) :- msw(init,Y0),msw(out(Y0),X0),hmm1(Y0,Xs,Ys).
       -- from unfolding (4) by (1)
 (7) hmm0([X0|Xs]) :- msw(init,Y0),msw(out(Y0),X0),hmm1(Y0,Xs).
       -- from folding (6) by (5)
 (8) hmm1(Y0,[]).
       -- unfolding (5) by (2) and (3) giving (8) and (9)
 (9) hmm1(Y0,[X|Xs]):- msw(tr(Y0),Y),msw(out(Y),X),hmm1(Y,Xs,Ys).
(10) hmm1(Y0,[X|Xs]):- msw(tr(Y0),Y),msw(out(Y),X),hmm1(Y,Xs).
       -- from folding (9) by (5)
\end{verbatim}
\rule{\textwidth}{0.25mm}\\ 
\caption{Unfold/fold program transformation of {\tt hmm0/1}}
\label{fig:dprism:trans}
\end{figure}

One  way   to  achieve   this  is  to   use  meaning   preserving  unfold/fold
transformation  for  logic  programs  \cite{Tamaki84,Pettorossi94}.  It  is  a
system of  program transformation containing  rules for unfolding  and folding
operations.  Unfolding replaces a goal with the matched body of a clause whose
head unifies  with the  goal and  folding is a  reverse operation.   There are
conditions  on   transformation  that   must  be  met   to  ensure   that  the
transformation is  meaning preserving,  i.e.\ the least  model of  programs is
preserved   through  transformation   (see   \cite{Tamaki84,Pettorossi94}  for
details).   Note that  meaning preserving  unfold/fold  program transformation
also  preserves the set  of all  explanations for  a goal.   So if  $\db_2$ is
obtained from $\db_1 \cup \{\mbox{\tt g(X):-g(X,Y)}\}$ by such transformation,
both programs have the same set  of all explanations for {\tt g(X)}, and hence
the  desired   property  ${\qdb}_1(\exists   {\tt  X}  \mbox{\tt   g(X,Y)})  =
{\qdb}_2(\mbox{\tt  g(X)})$ holds.   In  addition, usually,  $\db_2$ does  not
refer to the cumbersome ${\tt Y}$.

Fig.~\ref{fig:dprism:trans}   illustrates   a    process   of   such   program
transformation.   It  derives  an  HMM  program for  {\tt  hmm0(X)}  computing
incomplete data  from a  program for {\tt  hmm0(X,Y)} computing  complete data
using a  transformation system described  in \cite{Tamaki84}.  It  starts with
the initial program  defining {\tt hmm0(X,Y)} consisting of  \{{\tt (1)}, {\tt
  (2)}, {\tt  (3)}\} together  with two defining  clauses for  new predicates,
i.e.\ {\tt  (4)} for {\tt hmm0(X)}  and {\tt (5)} for  {\tt hmm1(Y0,Xs)}.  The
transformation process  begins by unfolding  the body goal {\tt  hmm0(X,Y)} in
{\tt (4)} by {\tt (1)} and  folding {\tt hmm1(Y0,Xs,Ys)} by {\tt (5)} follows,
resulting in {\tt  (7)}.  The defining clause {\tt  (5)} for {\tt hmm1(Y0,Xs)}
is processed similarly.  The final program obtained is \{{\tt (7)}, {\tt (8)},
{\tt  (10)}\}  which coincides  with  the HMM  program  for  {\tt hmm0(X)}  in
Fig.~\ref{prog:crf-hmm}.

This  example exemplifies  that unfold/fold  transformation has  the  power of
eliminating the redundant argument ${\tt  Y}$ in \mbox{\tt g(X):- g(X,Y)} and
deriving a specialized  program for ${\tt g(X)}$ that does  not refer to ${\tt
  Y}$ and hence  is suitable for tabling. However  how far this transformation
is generally applicable and how far it can be automated is future work.
%

\section{Discussion and future work}
\label{sec:discussion}

There are already  discriminative modeling languages for CRFs  such as Alchemy
\cite{Kok05}  based on  MLNs and  Factorie \cite{McCallum09}  based  on factor
graphs.   To define  potential functions  and  hence models,  the former  uses
weighted clauses  whereas the latter uses imperatively  defined factor graphs.
Both use Markov chain Monte-Carlo (MCMC) for probabilistic inference.  D-PRISM
differs from them in that although  programs are used to define CRFs like MLNs
and Factorie,  they are  purely generative, computing  output from  input, and
probabilities are computed exactly by dynamic programming.
%
TildeCRF  \cite{Gutmann06}  learns  CRFs   over  sequences  of  ground  atoms.
Potential  functions are computed  by weighted  sums of  relational regression
trees applied  to an  input sequence  with a fixed  size window.   TildeCRF is
purely  discriminative and  unlike  D-PRISM  uses a  fixed  type of  potential
function.  Also  it is  designed for linear-chain  CRFs and more  complex CRFs
such as CRF-CFGs are not intended or implemented. 

D-PRISM is interesting  from the viewpoint of statistical  machine learning in
that  it builds  discriminative models  from  generative models  and offers  a
general approach to implementing generative-discriminative pairs.  This unique
feature also makes it relatively easy and smooth to develop new discriminative
models     from     generative     models     as    we     demonstrated     in
Section~\ref{sec:new-models}.  In addition, as is shown by every experiment in
this paper,  there is  a clear trade-off  between accuracy  (by discriminative
models) and learning time (by generative  models), and hence we have to choose
which type  of model to  use, depending on  our purpose.  D-PRISM  assists our
choice by providing a unified environment to test both types.\\

Compared  to  PRISM,  D-PRISM has  no  restriction  on  programs such  as  the
uniqueness  condition,  exclusiveness  condition  and  independence  condition
\cite{Sato01g}.   Consequently non-exclusive  or  is permitted  in a  program.
Also probability computation  is allowed to fail by  constraints.  For example
it  is straightforward  to  add linguistic  constraints  such as  subject-verb
agreement  to a  PCFG  by adding  an  extra argument  carrying such  agreement
information to the  program.  Although loss of probability  mass occurs due to
disagreement in the generating  process, normalization recovers a distribution
and we  obtain a constraint  CRF-CFG as a  result.  Of course this  freedom is
realized  at the  cost of  normalization which  may be  prohibitive  even when
dynamic  programming is  possible.  This  would  happen when  adding too  many
constraints, e.g.,  agreement in number,  gender, tense and  so on to  a PCFG.
Thanks to the  removal of restrictive conditions however,  D-PRISM is now more
amenable to structure learning in ILP than PRISM, which is expected to open up
a new line of research of learning CRFs in ILP. \\

In  this paper  we concentrated  on learning  from complete  data in  CRFs and
missing value  is not considered. When  there are missing  values, for example
when some labels on a sequence in  a linear-chain CRF are missing, the data is
incomplete and parameter learning becomes much harder, if not impossible.
There is a  method of parameter learning from  incomplete data for conditional
distributions  using EM.   It is  developed  for PRISM  programs with  failure
\cite{Sato05}  and learns parameters  from a  conditional distribution  of the
form $\pdb(G_x \mid {\tt success})$ where  ${\tt success} = \exists x G_x$ and
$G_x$  is  a goal  for  incomplete  data $x$  that  may  fail.   The point  in
\cite{Sato05} is to automatically synthesize {\tt failure} predicate such that
$\pdb({\tt success})  = 1 -  \pdb({\tt failure})$ and rewrite  the conditional
distribution as an infinite series  $\pdb(G_x \mid {\tt success}) = \pdb(G_x)(
1 +  \pdb({\tt failure})  + \pdb({\tt  failure})^2 + \cdots)$  to which  EM is
applicable  (the  FAM   algorithm  \cite{Cussens01}).   Although  whether  the
adaptation of  this technique to EM  learning of CRFs with  incomplete data is
possible or not is unknown, it seems worth pursuing considering the simplicity
of EM compared to complicated gradient-based parameter learning algorithms for
incomplete data. \\

%
In Section~\ref{sec:trans}, the unfold/fold  program transformation is used to
remove the redundant argument ${\tt  Y}$ from ${\tt hmm0(X,Y)}$.  ${\tt Y}$ is
a   non-discriminating  argument  in   the  sense   of  \cite{Christiansen09}.
Christiansen and  Gallagher gave a  deterministic algorithm to  eliminate such
non-discriminating arguments without affecting the program's run-time behavior
\cite{Christiansen09}.   Actually deleting  non-discriminating  arguments from
clauses for  ${\tt hmm0(X,Y)}$ in Fig.~\ref{prog:crf-hmm} results  in the same
HMM program  obtained by program  transformation.  Compared to  their approach
however,  our  approach  is  based on  non-deterministic  unfold/fold  program
transformation and  allows for an introduction of  new predicates.  Clarifying
the relationship between these two approaches is future work. \\

Currently  only  binary features  or  their  counts  are allowed  in  D-PRISM.
Introducing real-valued features  is also a future work and  so is a mechanism
of parameter tying.
Finally, D-PRISM is experimentally implemented at the moment and we hope it is
part of the PRISM package in the future.

\section{Conclusion}
\label{sec:conclusion}
We   have  introduced   D-PRISM,   a  logic-based   generative  language   for
discriminative  modeling. As examples  show, D-PRISM  programs are  just PRISM
programs with  probabilities replaced  by weights.  It  is the  first modeling
language to our knowledge that  generatively defines CRFs and their extensions
to  probabilistic   grammars.   We  can  freely   build  logistic  regression,
linear-chain CRFs,  CRF-CFGs or new  models generatively with almost  the same
modeling cost  as PRISM while  achieving better performance  in discriminative
tasks.


\bibliographystyle{spmpsci}      


\end{document}